\documentclass[lettersize,journal]{IEEEtran}

\usepackage[bookmarks=true]{hyperref}
\hypersetup{
    colorlinks,
    linkcolor={red!50!black},
    citecolor={blue!50!black},
    urlcolor={blue!80!black}
}
\usepackage{amsmath,amsfonts}
\usepackage{algorithm}
\usepackage{algpseudocode}
\usepackage{amsmath}
\usepackage{array}
\usepackage{textcomp}
\usepackage{stfloats}
\usepackage{url}
\usepackage{verbatim}
\usepackage{graphicx}
\usepackage{cite}
\usepackage{xcolor}
\usepackage{siunitx}
\usepackage{subcaption}
\usepackage{siunitx}
\usepackage{array}
\usepackage{enumitem} 
\usepackage{caption}
\usepackage{lipsum} 
\usepackage{flushend}

\usepackage{multirow}
\usepackage{longtable}
\usepackage{ragged2e} 
\usepackage{makecell}
\usepackage{booktabs}

\DeclareSIUnit\cell{cell}
\DeclareSIUnit\cells{cells}
\DeclareSIUnit\trees{trees}

\usepackage{listings}
\usepackage{xcolor}

\lstdefinelanguage{ROSmsg}{
  morekeywords={float32, int8, bool}, 
  sensitive=true,
  morecomment=[l]{\#}, 
  morestring=[b]",
}

\usepackage[justification=centering]{caption} 

\begin{document}

\title{Minimalistic Autonomous Stack for High-Speed Time-Trial Racing}

\author{Mahmoud Ali, Hassan Jardali, Youwei Yu, Durgakant Pushp, Lantao Liu
\thanks{The authors are with the Luddy School of Informatics,
Computing, and Engineering at Indiana University, Bloomington, IN 47408,
USA. E-mail: \texttt{\{alimaa, hjardali, youwyu, dpushp, lantao\}@iu.edu}}
}
\markboth{Journal of \LaTeX\ Class Files,~Vol.~14, No.~8, August~2021}%
{Shell \MakeLowercase{\textit{et al.}}: A Sample Article Using IEEEtran.cls for IEEE Journals}
\maketitle
\begin{abstract}
Autonomous racing has seen significant advancements, driven by competitions such as the Indy Autonomous Challenge (IAC) and the Abu Dhabi Autonomous Racing League (A2RL). However, developing an autonomous racing stack for a full-scale car is often constrained by limited access to dedicated test tracks, restricting opportunities for real-world validation. 
While previous work typically requires extended development cycles and significant track time, this paper introduces a minimalistic autonomous racing stack for high-speed time-trial racing that emphasizes rapid deployment and efficient system integration with minimal on-track testing.
The proposed stack was validated on real speedways, achieving a top speed of \SI{206}{\kilo\meter\per\hour} within just 11 hours' practice run on the track with \SI{325}{\kilo\meter} in total. Additionally, we present the system performance analysis, including tracking accuracy, vehicle dynamics, and safety considerations, offering insights for teams seeking to rapidly develop and deploy an autonomous racing stack with limited track access.
\end{abstract}

\begin{IEEEkeywords}
Autonomous Systems, Autonomous Racing, Resilience System
\end{IEEEkeywords}

\vspace{-16pt}
\section{Introduction}
 Autonomous racing has gained significant traction in recent years, advancing both research and real-world deployment in high-speed autonomy. Competitions such as
 IAC~\cite{indyautonomouschallengeIndyAutonomous} and A2RL~\cite{a2rlDhabiAutonomous} provide a platform for testing cutting-edge autonomous systems in extreme conditions. These events have driven advancements in perception, planning, and control algorithms~\cite{saba2024fast, raji2024er, betz2023tum, chung2024autonomous}, leading to fully autonomous race cars competing at \SI{290}{\kilo\meter\per\hour}~\cite{top_results}.
Despite these achievements, developing an autonomous racing stack for full-scale vehicles remains a resource-intensive endeavor due to the limited availability of dedicated racetracks for testing. Prior work has demonstrated impressive results but often relies on years of development and extensive track-testing time, making rapid deployment difficult for new teams. 

To address this challenge, we present a minimalistic autonomous racing stack designed for high-speed time-trial racing with a focus on single-car speed performance and rapid deployment. 
Our approach strategically maximizes track time utilization, enabling a fully functional autonomy stack with minimal on-track testing. 
The proposed system was implemented on the IAC AV-24 race car~\cite{AV24} and validated on real speedways, achieving a top speed of \SI{206}{\kilo\meter\per\hour} with only 11 track hours and \SI{325}{\kilo\meter} of practice runs.
The key contributions of this work include:
I) A minimalistic autonomous racing stack tailored for high-speed solo racing. 
II) A system integration of the proposed stack into the AV-$24$ racing car. 
III) An evaluation of the proposed stack
with emphasis on controllers performance and dynamics analysis. 
IV) A discussion of the safety measures incorporated into the system, along with an analysis of failure cases.
\begin{figure}[t!]\vspace{-5pt}
{
    \centering
    \includegraphics[scale=0.055]{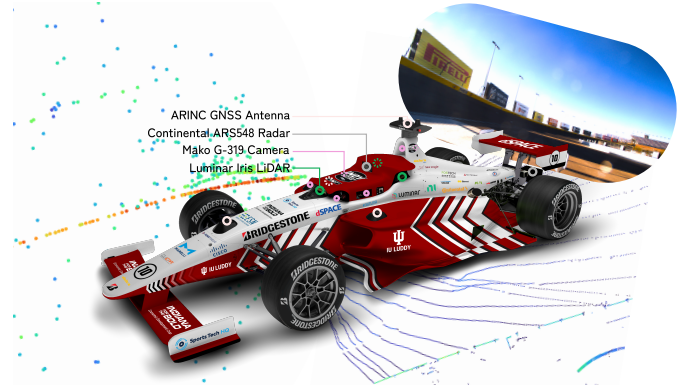} \vspace{-10pt}
    \caption{\small IAC AV24 race car features a sensor-rich system to attain high-speed autonomy.  \vspace{-10pt}}
    \label{fig:race_car}
}
\end{figure}
\vspace{-12pt}
\section{System}
In this section, we introduce the IAC AV-24 racing car, its sensor suite, and the actuation system, which serves as the primary interface to the car.
\vspace{-12pt}
\subsection{Hardware Overview}
The IAC AV24 autonomous racing car is based on the Dallara IL-15 employed in the Indy Lights Racing Series. The AV24 features a \SI{2.0}{\liter} single-turbocharged, inline $4$-cylinder engine with a maximum power of $488$ HP,
paired with a six-speed sequential gearbox.
The car is converted to be autonomous by replacing the cockpit with autonomous driving components, including a computing platform, sensors, an actuation system, and communication apparatuses.
The vehicle’s actuation is enabled through a Schaeffler Paravan ``SpaceDrive II" drive-by-wire (DBW) system.

The AV24 is equipped with a {dSpace AUTERA} 
AutoBox PC, featuring an Intel Xeon CPU, an Nvidia A5000 GPU, and 14 TB of storage capacity. The AV24 is also equipped with a localization suite comprising three Global Navigation Satellite Systems (GNSS): two {Novatel PwrPak7D}
(dubbed Novatel-A and Novatel-B) and one {VectorNav-310}.
The AV24 perception stack includes six cameras—two front-facing, long-range cameras configured as a stereo setup and four wide Field-of-View (FoV) cameras providing \num{360}$^\circ$ coverage—along with two {Continental ARS540}
4D radars (one front and one rear), each featuring a \num{60}$^\circ$ FoV, as well as three {Luminar Iris}
LiDARs, each with a \num{120}$^\circ$ horizontal FoV to ensure complete \num{360}$^\circ$ coverage. Fig.~\ref{fig:race_car} shows our race car with various sensors with real data from exteroceptive sensors, including the front Radar, left LiDAR, and rear camera.
\begin{figure}[h]
    \centering
    \includegraphics[scale=1.0]{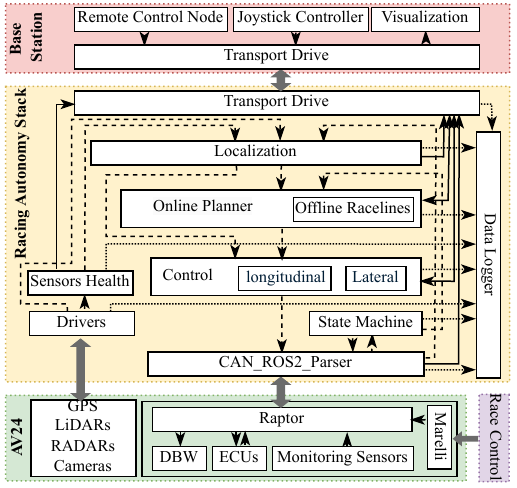} \vspace{-5pt}
    \caption{\small System diagram.}
    \label{fig:system_diagram}
    \vspace{-0.25in}
\end{figure}
\vspace{-12pt}
\subsection{Actuation Interface Systems}
The actuation system at a lower level comprises an embedded platform that includes Electronic Control Units (ECUs), servomotors for steering, throttle, and braking, a voltage management system, and a suite of monitoring sensors measuring values such as engine oil temperature and pressure, coolant temperature, tire temperature and pressure, engine RPM, wheel speed, wheel torque, and battery voltage. The New Eagle GCM 196 Raptor is used as the interface to this lower-level system, simultaneously handling the transmission of throttle, brake, gear-shift, and steering commands to the DBW system and routing sensor data to the autonomy stack. This interfacing is accomplished through the Controller Area Network (CAN) serial bus system.

The lower-level software implements a state machine that governs the vehicle’s operational status throughout its run. This state machine transitions between states by verifying key functionalities, such as performing an actuation test, to ensure proper steering and braking. Additionally, it continuously monitors communication between the Raptor and the onboard computer via a rolling counter mechanism. 
Once all functionality tests have been passed, the engine can be started, allowing the vehicle to move into higher-level states (e.g., driving mode) and begin receiving commands from the autonomy stack. In the event of anomalies (e.g., a lost connection between the Raptor and the autonomy stack), the system initiates a supervised stop or an emergency shutdown.

Another component interfacing with the lower-level system is Race Control, serving as a supervisory entity responsible for managing the race by issuing command flags that instruct the vehicle to stop, adjust speed, or return to the pit. Two categories of flags are employed: vehicle flags, which direct an individual car to start its engine, enter the pit lane, or come to a complete stop; and track flags, which are broadcast to all cars and, e.g., define the maximum allowable speed. These flags are transmitted to AV24 via a Marelli-powered system.
\vspace{-8pt}
\section{Minimalistic Autonomy Stack}
We describe our approach to implementing a C++/ROS2-based, minimalistic, efficient, and safety-critical autonomy stack for high-speed time-trial (single-car) racing.
We also discuss the challenges encountered, including sensor driver integration,
adherence to execution deadlines in estimation, planning, and control modules, as well as data recording, visualization, and communication between the stack and both the lower-level system and the base station.

\vspace{-8pt}
\subsection{Autonomy Stack Peripherals}
\subsubsection{Raptor to Autonomy Stack Communication}
A DBC (Database CAN) file provided by the competition organizers is used to interpret the CAN messages. To integrate these messages within the stack, we 
leverage the raptor can driver~\cite{can_dbc}, to convert the CAN messages coming from raptor into ROS messages and publish them on multiple topics, as well as convert control commads from ROS messages to CAN messages to send it to raptor.

\subsubsection{Autonomy Stack to Base Station Communication}
\label{stack2base_comm}
We use Secure Shell (SSH) to remote access the onboard computer to launch our stack. However, rather than relying on the ROS2 framework for data exchange between the onboard PC and the base station, we use the transport driver package~\cite{transport_drivers}
in conjunction with the ASIO library~\cite{asio}. This approach enables us to serialize and encode only the essential data, thereby reducing network load and improving latency. 
The transmitters and receivers on the base station run at a frequency of 10 Hz. 
\subsubsection{Sensor Drivers}
To bridge the barrier from the systematic complexity to the agile development, we leverage VLANs out of the box to broadcast different modules into communicable domains. All sensors use official C\texttt{++} drivers with Unix system time consistently, while GNSS systems stay with the GPS reference time. Here we detail the specifications: (1) three LiDARs run at \SI{10}{\hertz} with trapezoidal scan patterns, (2) six cameras run with $ 2064 \times 1544 $ resolution at \SI{10}{\hertz} with automatic exposure and amplification, (3) two NovAtels measure position and heading at \SI{20}{\hertz} and IMU at \SI{125}{\hertz}, (4) one VectorNav measures position and heading at \SI{20}{\hertz} and IMU at \SI{800}{\hertz}. For each GNSS system, we have a RTK subscription to achieve off-the-shelf real-time positioning in the East-North-Up (ENU) frame. Instead of compiling all nodes into one single ROS-Launch file, which is convenient but may lead to performance loss with dropped messages, we create a standalone manager to control sensors and monitor the hardware functionality.

\vspace{-8pt}
\subsection{Vehicle State Estimation}
Our state estimation (localization), based on Error State Kalman Filter (ESKF), enjoys the RTK precision, but with compromise in robustness. To tackle this, we propose three takeaways: (1) raw message reliability check with multiple validations, (2) dead-reckoning timeout limits and robust re-initialization, and (3) outlier-robust Kalman filtering through generalized Bayes. First, to validate the raw message reliability, we consider the GNSS-estimated variance, RTK solution status, and motion prior via history states. Second, during unexpected RTK solution failures, we must rely on other sensors (e.g., IMU), but as time exceeds the threshold we reinitialize the ESKF system. This can only be achieved under minimal localization shifts dynamically w.r.t. current velocity. Third, the inverse multi-quadratic weighting function~\cite{pmlr-v235-duran-martin24a} serves to further mitigate the algorithmic outliers comparing to our previous safety-critical checks. Note that we use RTK position to estimate the horizontal velocity and heading, while we only rely on IMU for angular velocity because its linear acceleration measurements are corrupted by the vehicle vibrations. Our ESKF primarily uses NovAtel-A and outputs localization at \SI{125}{\hertz}.

Towards the general 3D localization, two GNSS systems are coupled with ESKF loosely. The RTK fused with two-IMU localization~\cite{10161187} gives GPS-Inertial odometry a precise initial value, and in turn, iSAM2~\cite{isam2} optimizes the roll and pitch angles estimation at which the single GNSS usually underperforms due to the height measurement noise and partial observability. Note that we localize both systems simultaneously in the ENU frame and use the installation specification as default extrinsic coefficients. Additionally, we treat the racing track banking data~\cite{Rowold2023IV} as observations to further correct the roll and pitch in the ESKF system.

\vspace{-8pt}
\subsection{Path Planners}
Our planning module includes two planners: 
(1) {\em Offline Global Planner: }
For high-speed solo laps, the raceline (i.e., the reference path) is computed offline and subsequently provided to the online planner during the run. To generate this raceline, the track boundaries are first extracted from Google Earth \cite{googleGoogleEarth} as a KML file. After smoothing these boundaries, the raceline is generated using the minimum-curvature method \cite{heilmeier2020minimum, githubGitHubTUMFTMglobal_racetrajectory_optimization}. The raceline is defined by the variables \textit{x}, \textit{y}, and \textit{v\_ref}, where \textit{x} and \textit{y} are the reference coordinates in the ENU frame, and \textit{v\_ref} is the maximum allowed longitudinal velocity at each point of the raceline. The generated racelines are saved as csv files. 
(2) {\em Online Planner:}
The online planner loads the offline-generated racelines, represented as quintic splines. It begins by taking the vehicle’s current pose from the state estimation module and determining the nearest point on the raceline via the Newton's method.
Next, the planner generates the path points based on the vehicle’s current velocity, a defined time horizon, and a specified step size. Once the path is determined, it is transformed from the global ENU coordinate frame to the car’s local frame to be used by the controller. 

The planner also integrates commands from the base station, sent through the remote-control node, and responds to flags from race control. The remote-control node may adjust the maximum velocity or switch the active raceline if needed. In turn, the planner enforces velocity limits based on these directives; for example, if race control issues a yellow flag, the planner will cap \textit{v\_ref} at 80 mph whenever the computed reference velocity exceeds that threshold.

\vspace{-8pt}
\subsection{Controllers}
The control commands for operating the vehicle are categorized into \textit{longitudinal} (throttle, brake, and gear) and \textit{lateral} (steering) commands. 
\subsubsection{Longitudinal Controller} A PID controller is utilized, with the reference velocity defined as the velocity of the \textit{the second} point from the locally planned trajectory. This method is employed to mitigate the effects of delays within the system.
If \(\Delta v = v_{\text{ref}} - v_{\text{car}}\) is positive and exceeds a specified throttle deadband, throttle is applied. Conversely, if \(\Delta v\) is sufficiently negative (i.e., it falls below a brake deadband threshold), the brake PID is activated to reduce speed. Engine gear shifting is also handled by the same controller, using engine RPM thresholds to determine when to shift up or down.
\subsubsection{Lateral Controller} The Pure Pursuit algorithm~\cite{coulter1992implementation} is implemented. This method determines the desired steering angle \(\delta\) based on a short-horizon path that connects the vehicle’s current position to a look-ahead point on the planned trajectory. 
The steering command is computed using the following equation:  
$
\delta = \tan^{-1} \left( \frac{2L \sin{\alpha}}{L_d} \right)
$,
where \(L\) is the vehicle’s wheelbase, \(\alpha\) is the heading error (the angle between the vehicle’s heading and the line connecting the vehicle to the look-ahead point), and the adaptive look-ahead distance
$
L_d = \min(L_{\mathrm{d,min}} + k \cdot v_{\text{car}},\, L_{\mathrm{d,max}}),  
$
where \(k\) is a proportional scaling factor. Finally, the computed command \(\delta\) is multiplied by the steering ratio to produce the actual wheel angle.

\subsubsection{Remote and Joystick Controller}
A remote control node runs on the base station PC to transmit flags, maximum speed settings, and target racelines to the vehicle. 
Under the competition rules, incremental maximum speed commands may be issued before reaching the start-finish line. 
Additionally, a joystick node on the base station enables manual override of controller commands to protect the car in emergencies.
\subsubsection{Safety Considerations}  
To ensure safe operation on the track, we implemented multiple safety mechanisms in the stack. Timeouts are used to monitor module connectivity via heartbeat messages, and failures trigger appropriate safety actions; localization failure applies brakes while maintaining the last steering command, while a planner crash leads to a controlled stop along the global trajectory. A cross-track error threshold ensures the car stops if deviation exceeds a predefined limit to prevent going off-track. Sensor health monitoring continuously assesses sensor functionality, stopping the vehicle if critical failures occur. 
A supervisory state machine monitors the lower-level state machine and integrates additional inputs from the stack to modify the vehicle state.
Command validation ensures executed actions match intended commands through hardware feedback. 
\begin{figure*}[ht]
    \vspace{-0.1in}
    \centering
    \begin{subfigure}{\linewidth}
        \centering
        \includegraphics[width=\linewidth, trim=0 0 0 53, clip]{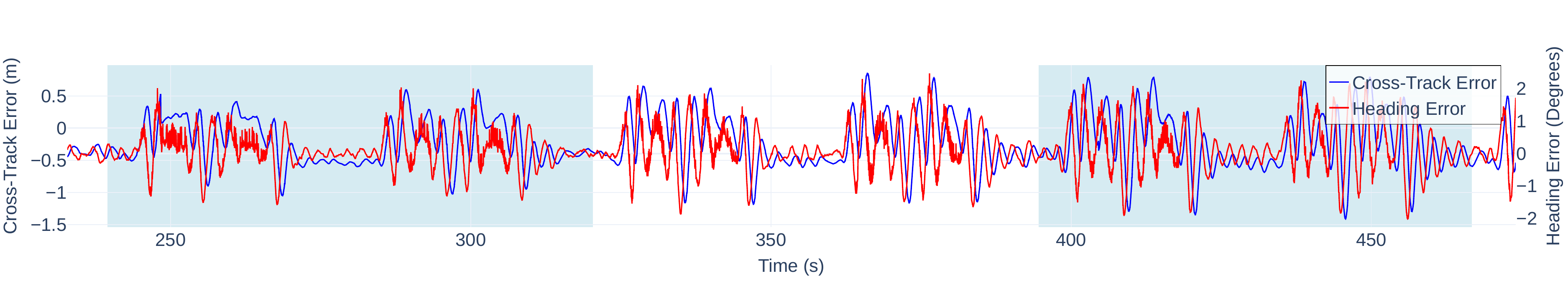}\vspace{-5pt}
        \caption{Cross-track and heading errors.}
        \label{fig:errors_vs_time}
    \end{subfigure}
    \vspace{-0.1in}
    \begin{subfigure}{\linewidth}
        \centering
        \includegraphics[width=\linewidth, trim=0 0 0 65, clip]{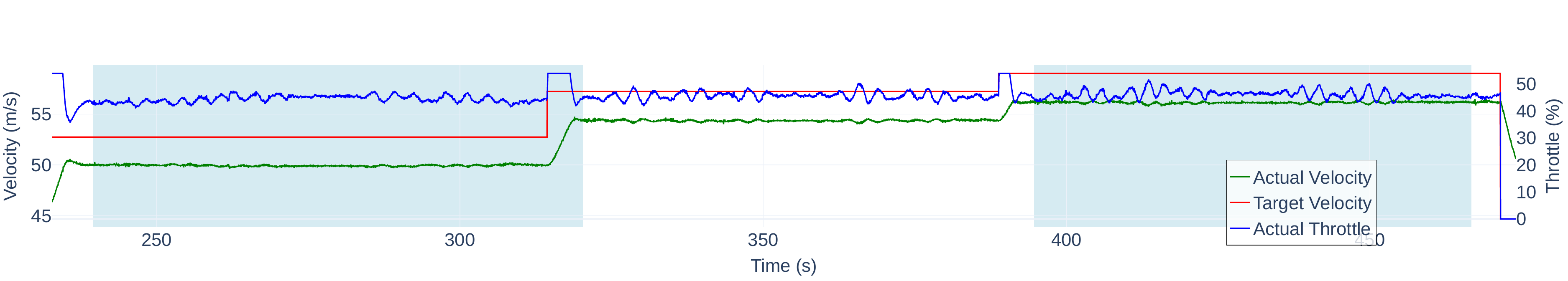}
        \caption{Target velocity, actual velocity, and applied throttle.}
        \label{fig:velocity_throttle_vs_time}
    \end{subfigure}
    \caption{\small Tracking errors during the IAC competition. Three laps are shown above, each distinguished by a different background color.}
    \label{fig:tracking_errors}
    \vspace{-0.2in}
\end{figure*}

\vspace{-8pt}
\subsection{Data Visualization and Logging}
\subsubsection{Visualization}
The base station receives critical data for real-time performance monitoring. 
These data are converted into ROS messages and visualized using Foxglove~\cite{foxglove}. A 3D panel displays track borders, reference racelines, estimated localization, and planned local trajectory, providing a qualitative assessment of vehicle performance. Additional time series plots and raw messages (e.g., cross-track error, heading error, throttle, steering, target/current velocities, and vehicle/track flags) offer a quantitative evaluation. 
This visualization introduces a second layer of safety by enabling a manual triggering of a safe stop if any metric deviates from its acceptable range.
\subsubsection{Data recording}
MCAP, configured with \SI{1}{\giga\byte} mini-chunks, is used to record all ROS-related data. Meanwhile, separate scripts are executed to log system states, including CPU usage, memory consumption, network activity, system load, process queue, running Linux processes, PCI device status, CAN bus activity, and system journal logs. To balance lossless data recording with CPU efficiency, different data sources can be assigned to separate threads as needed. Additionally, clearing the system cache after completing the recording is recommended to free up memory. In our final run, the bags recorded without the perception sensors reached a total size of \SI{3.7}{\giga\byte}.
\begin{figure}[ht] \vspace{-0.05in}
    \centering
        \includegraphics[height=1.1in]{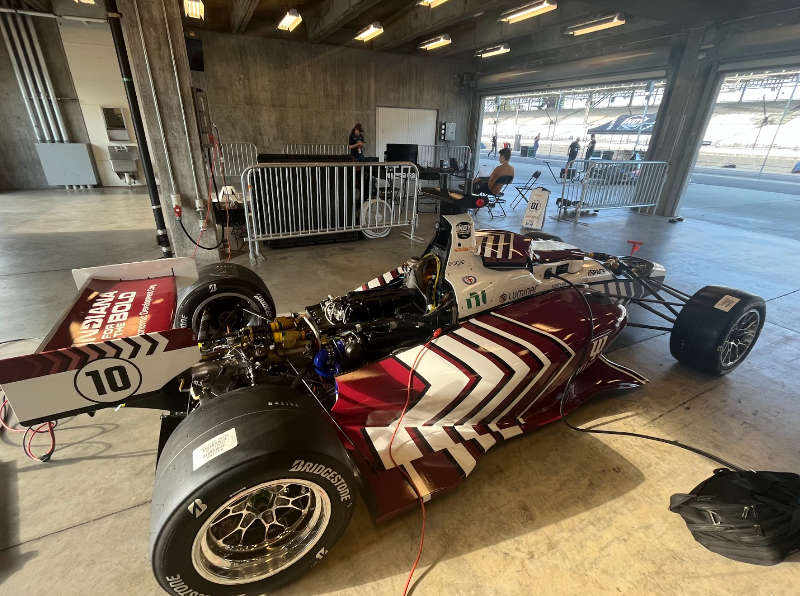} \ 
        \includegraphics[height=1.1in]{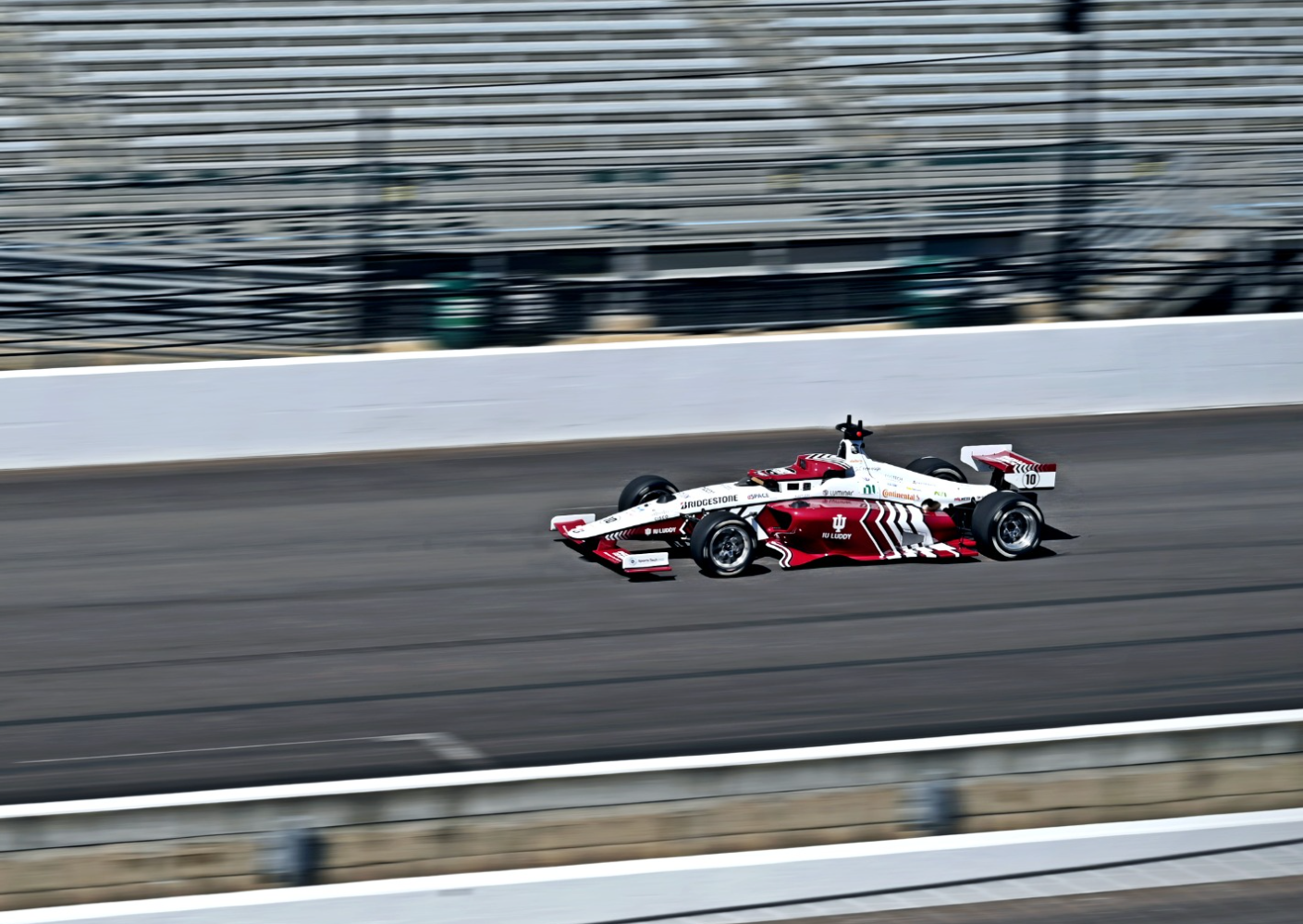}
    \caption{\small Left: Final car testing before the race. Right: car in competition at IMS. }
    \label{fig:car-in-track}
    \vspace{-0.15in}
\end{figure}
\vspace{-4pt}
\section{Results}
We validated our stack using AWSIM~\cite{autonomalabsOverviewAWSIM}, a Unity-based simulator that closely replicates the control interface to the real car.
However, in this section, we present only the real-world results from our run at IAC in the Indianapolis Motor Speedway (IMS) on September 6, 2024 (see Fig.~\ref{fig:car-in-track}), where a lap-average speed of \SI{200}{\kilo\meter\per\hour} was achieved; the run was live-streamed on Youtube; check \href{https://youtu.be/UitZRC3KpNE}{https://youtu.be/UitZRC3KpNE}. 

\subsubsection{Tracking Errors}
Figure~\ref{fig:errors_vs_time} illustrates the performance of the lateral controller, revealing that both heading and cross-track errors increase at higher velocities shown in Fig. \ref{fig:velocity_throttle_vs_time}. 
Specifically, the cross-track error range expanded from $[-1, 0.6]$\si{\meter} during the first lap (at \SI{50}{\meter\per\second}) to $[-1.4, 0.8]$\si{\meter} during the third lap (at \SI{56}{\meter\per\second}).
Similarly, the heading error range expanded from $[-1.5, 2.0]$\si{\degree} in the first lap to $[-2.0, 2.25]$\si{\degree} in the third lap.
In contrast, Figure~\ref{fig:velocity_throttle_vs_time} shows that the longitudinal PID velocity controller maintains a consistent error of approximately \SI{2.5}{\meter\per\second}. 
Notably, the system oscillations are primarily due to the pure pursuit controller's continuous targeting of a fixed velocity.
\subsubsection{Dynamics}
\begin{figure}[ht]
    \vspace{-0.1in}
    \centering
    \begin{subfigure}[b]{0.49\linewidth}
        \centering
        \includegraphics[scale=0.95]{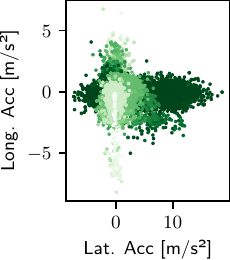}
        \caption{G-G diagram}
        \label{fig:gg_diagram-label}
    \end{subfigure}
    \hfill
    \begin{subfigure}[b]{0.49\linewidth}
        \centering
        \includegraphics[scale=0.95]{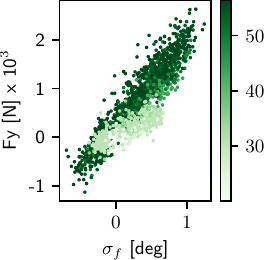}
        \caption{$F_y$ Vs $\sigma_{f}$}
        \label{fig:dynamics-label}
    \end{subfigure} 
    \caption{\small Left: G-G diagram; Right: Measured $F_y$ vs. front slip angle $\sigma_f$ with speed indicated by color [\SI{}{\meter\per\second}].}
    \label{fig:tracking_errors}
    \vspace{-0.15in}
\end{figure}

The G-G diagram in Fig.~\ref{fig:gg_diagram-label} shows that the car reached a maximum lateral acceleration of \SI{18}{\meter\per\second\squared} according to IMU data. 
Lateral force versus slip angle curves (computed using \cite{becker2023model}) show that the tires operate in the linear range of the Pacejka model (Fig.~\ref{fig:dynamics-label}). These values serve as initial inputs for model-based controllers—either as constraints with maximum acceleration limits \cite{wischnewski2022tube} or for extracting cornering stiffness \cite{raji2024er}.

\vspace{-4pt}
\section{Discussion}
\label{sec:discussion}
One limitation of the current stack is that the current localization module was unable to tightly couple data from multiple GNSS receivers due to inconsistent motion shifts. Although the current stack can leverage auxiliary sensor motion estimation to improve accuracy,  the dependence on the single main sensor will lead to safety-critical issues due to lack of a smooth transition. Better tuning of the controllers' parameters could improve velocity and lateral tracking. In addition, improving controller's stability performance will require an advanced controller that incorporates vehicle dynamics model.

We encountered one system failure that caused the ROS2 middleware to intermittently freeze, delaying the rolling counter sent to the raptor. Consequently, the vehicle entered emergency mode, applying the brakes and locking the last steering command, as it exited the second turn at IMS, resulting in a crash.
Additional details are in the supplementary material.
\vspace{-12pt}
\section{Conclusion}
We presented a minimalistic autonomous racing stack that can be rapidly deployed and effectively validated with limited track time. Our approach achieved a top speed of \SI{206}{\kilo\meter\per\hour} after only \num{11} hours of practice runs.
The results obtained from this stack are valuable for subsequent development iterations, serving to tune model-based controllers as well as other safety, perception, and planning parameters.
\newpage

\renewcommand{\arraystretch}{1.3} 
\setlength{\tabcolsep}{10pt}      
\vspace{50pt}
\begin{table*}
\centering
\begin{tabular}{|p{0.28\linewidth}|p{0.68\linewidth}|}
\multicolumn{2}{c}{{Supplementary Materials Description}} \\ 
\hline
\textbf{Supplementary Material} & \textbf{Description} \\
\hline
\href{https://doi.org/10.5281/zenodo.17187680}{https://doi.org/10.5281/zenodo.17187680}
& The ROS bag recorded during the IAC competition at IMS is available via the attached link. It encompasses all the data visualized in the results section, in addition to raw GPS data, state estimation, planning, control, and visualization messages. The dataset further includes lower-level monitoring sensors' readings. Additionally, we have provided our message definition package to facilitate accurate data interpretation.\\
\hline
\textbf{supplementary\_material.pdf} & This document comprises the pseudocode for the online planner as well as the lateral and longitudinal controllers, along with the ROS messages definitions and the parameters used during the IAC competition. It also visualizes and explains the CAN logs associated with the system failure and crash encountered at IMS before the IAC competiton, providing insight into the incident dynamics. \\
\hline
\href{https://youtu.be/-yaAK8FA0qs}{https://youtu.be/-yaAK8FA0qs} & The video presents the live-streamed footage of the car on the track during the competition, accompanied by the base station visualizations of all critical performance metrics required for monitoring the car performance. \\
\hline
\href{https://youtu.be/6Vt7Q4ySETQ}{https://youtu.be/6Vt7Q4ySETQ}& The video shows various experimental setups utilized while testing our stack, including static testing in the garage/pit and actual testing on the track, both with and without a chase car.\\
\hline
\end{tabular}
\end{table*}

\section{Acknowledgments}
We wish to acknowledge all the students and researchers who advanced the software stack and assisted during practice sessions, as well as the technical and logistics personnel of the Indy Autonomous Challenge for their dedicated support.


\bibliographystyle{IEEEtran}
\bibliography{references}            

\begin{thebibliography}{10}
\providecommand{\url}[1]{#1}
\csname url@samestyle\endcsname
\providecommand{\newblock}{\relax}
\providecommand{\bibinfo}[2]{#2}
\providecommand{\BIBentrySTDinterwordspacing}{\spaceskip=0pt\relax}
\providecommand{\BIBentryALTinterwordstretchfactor}{4}
\providecommand{\BIBentryALTinterwordspacing}{\spaceskip=\fontdimen2\font plus
\BIBentryALTinterwordstretchfactor\fontdimen3\font minus \fontdimen4\font\relax}
\providecommand{\BIBforeignlanguage}[2]{{%
\expandafter\ifx\csname l@#1\endcsname\relax
\typeout{** WARNING: IEEEtran.bst: No hyphenation pattern has been}%
\typeout{** loaded for the language `#1'. Using the pattern for}%
\typeout{** the default language instead.}%
\else
\language=\csname l@#1\endcsname
\fi
#2}}
\providecommand{\BIBdecl}{\relax}
\BIBdecl

\bibitem{indyautonomouschallengeIndyAutonomous}
``{I}ndy {A}utonomous {C}hallenge --- indyautonomouschallenge.com.''

\bibitem{a2rlDhabiAutonomous}
``{A}bu {D}habi {A}utonomous {R}acing {L}eague in {U}{A}{E} | {A}2{R}{L} --- a2rl.io.''

\bibitem{saba2024fast}
A.~Saba, A.~Adetunji, A.~Johnson, A.~Kothari, M.~Sivaprakasam, J.~Spisak, P.~Bharatia, A.~Chauhan, B.~Duff~Jr, N.~Gasparro \emph{et~al.}, ``Fast and modular autonomy software for autonomous racing vehicles,'' \emph{arXiv preprint arXiv:2408.15425}, 2024.

\bibitem{raji2024er}
A.~Raji, D.~Caporale, F.~Gatti, A.~Toschi, N.~Musiu, M.~Verucchi, F.~Prignoli, D.~Malatesta, A.~F. Jesus, A.~Finazzi \emph{et~al.}, ``er. autopilot 1.1: A software stack for autonomous racing on oval and road course tracks,'' \emph{IEEE Transactions on Field Robotics}, 2024.

\bibitem{betz2023tum}
J.~Betz, T.~Betz, F.~Fent, M.~Geisslinger, A.~Heilmeier, L.~Hermansdorfer, T.~Herrmann, S.~Huch, P.~Karle, M.~Lienkamp \emph{et~al.}, ``Tum autonomous motorsport: An autonomous racing software for the indy autonomous challenge,'' \emph{Journal of Field Robotics}, vol.~40, no.~4, pp. 783--809, 2023.

\bibitem{chung2024autonomous}
C.~C. Chung, A.~Finazzi, H.~Seong, D.~Lee, S.~Lee, B.~Kim, G.~Gang, and D.~H. Shim, ``Autonomous system for head-to-head race: Design, implementation and analysis; team kaist at the indy autonomous challenge,'' \emph{IEEE Transactions on Field Robotics}, 2024.

\bibitem{top_results}
``{I}ndy {A}utonomous {C}hallenge ({IAC}) media --- https://www.indyautonomouschallenge.com/ims-results-2024.''

\bibitem{AV24}
``Dallara {AV}-24 --- https://www.indyautonomouschallenge.com/racecar.''

\bibitem{can_dbc}
``{DBW} {CAN} driver --- https://github.com/neweagleraptor/raptor-dbw-ros2/.''

\bibitem{transport_drivers}
``Transport drivers --- https://github.com/ros-drivers/transport\_drivers.''

\bibitem{asio}
``{AISO} library --- https://think-async.com/asio/.''

\bibitem{pmlr-v235-duran-martin24a}
\BIBentryALTinterwordspacing
G.~Duran-Martin, M.~Altamirano, A.~Shestopaloff, L.~S\'{a}nchez-Betancourt, J.~Knoblauch, M.~Jones, F.-X. Briol, and K.~P. Murphy, ``Outlier-robust kalman filtering through generalised {B}ayes,'' in \emph{Proceedings of the 41st International Conference on Machine Learning}, ser. Proceedings of Machine Learning Research, R.~Salakhutdinov, Z.~Kolter, K.~Heller, A.~Weller, N.~Oliver, J.~Scarlett, and F.~Berkenkamp, Eds., vol. 235.\hskip 1em plus 0.5em minus 0.4em\relax PMLR, 21--27 Jul 2024, pp. 12\,138--12\,171. [Online]. Available: \url{https://proceedings.mlr.press/v235/duran-martin24a.html}
\BIBentrySTDinterwordspacing

\bibitem{10161187}
Y.~Yu, Y.~Liu, F.~Fu, S.~He, D.~Zhu, L.~Wang, X.~Zhang, and J.~Li, ``Fast extrinsic calibration for multiple inertial measurement units in visual-inertial system,'' in \emph{2023 IEEE International Conference on Robotics and Automation (ICRA)}, 2023, pp. 01--07.

\bibitem{isam2}
M.~Kaess, H.~Johannsson, R.~Roberts, V.~Ila, J.~Leonard, and F.~Dellaert, ``isam2: Incremental smoothing and mapping with fluid relinearization and incremental variable reordering,'' in \emph{2011 IEEE International Conference on Robotics and Automation}, 2011, pp. 3281--3288.

\bibitem{Rowold2023IV}
M.~Rowold, L.~Ögretmen, U.~Kasolowsky, and B.~Lohmann, ``Online time-optimal trajectory planning on three-dimensional race tracks,'' in \emph{2023 IEEE Intelligent Vehicles Symposium (IV)}, 2023, pp. 1--8.

\bibitem{googleGoogleEarth}
``{G}oogle {E}arth --- earth.google.com,'' \url{https://earth.google.com/web/}, [Accessed 14-02-2025].

\bibitem{heilmeier2020minimum}
A.~Heilmeier, A.~Wischnewski, L.~Hermansdorfer, J.~Betz, M.~Lienkamp, and B.~Lohmann, ``Minimum curvature trajectory planning and control for an autonomous race car,'' \emph{Vehicle System Dynamics}, 2020.

\bibitem{githubGitHubTUMFTMglobal_racetrajectory_optimization}
``{G}it{H}ub - {T}{U}{M}{F}{T}{M}/global\_racetrajectory\_optimization: {T}his repository contains multiple approaches for generating global racetrajectories. --- github.com,'' \url{https://github.com/TUMFTM/global_racetrajectory_optimization}, [Accessed 14-02-2025].

\bibitem{coulter1992implementation}
R.~Coulter, ``Implementation of the pure pursuit path tracking algorithm,'' 1992.

\bibitem{foxglove}
``Foxglove --- https://foxglove.dev/.''

\bibitem{autonomalabsOverviewAWSIM}
``{O}verview - {A}{W}{S}{I}{M} racing document --- autonomalabs.github.io,'' \url{https://autonomalabs.github.io/AWSIM/}, [Accessed 16-02-2025].

\bibitem{becker2023model}
J.~Becker, N.~Imholz, L.~Schwarzenbach, E.~Ghignone, N.~Baumann, and M.~Magno, ``Model-and acceleration-based pursuit controller for high-performance autonomous racing,'' in \emph{2023 IEEE International Conference on Robotics and Automation (ICRA)}.\hskip 1em plus 0.5em minus 0.4em\relax IEEE, 2023, pp. 5276--5283.

\bibitem{wischnewski2022tube}
A.~Wischnewski, T.~Herrmann, F.~Werner, and B.~Lohmann, ``A tube-mpc approach to autonomous multi-vehicle racing on high-speed ovals,'' \emph{IEEE Transactions on Intelligent Vehicles}, vol.~8, no.~1, pp. 368--378, 2022.

\end{thebibliography}

\clearpage
\onecolumn


\lstdefinelanguage{ROSmsg}{
  morekeywords={float32, int8, bool}, 
  sensitive=true,
  morecomment=[l]{\#}, 
  morestring=[b]",
}

\lstset{
  language=ROSmsg,
  basicstyle=\ttfamily\footnotesize, 
  keywordstyle=\color{blue}, 
  commentstyle=\color{gray}, 
  stringstyle=\color{red}, 
  breaklines=true,
  frame=single, 
  captionpos=b
}









\vspace{-1in}
\section*{\textbf{Supplementary Material}: Pseudocode, parameters, ROS Messages, and System Failure}
\subsection{Pseudocode}

\begin{algorithm}[H]
\caption{Online Planner}
\label{alg:online_planner_timer}
\begin{algorithmic}[1]
\Require Offline-generated \(\textit{racelines}\), time horizon \(\Delta T\), step size \(\delta t\)
\State \textbf{Load} \(\textit{racelines}\) from files
\State \textbf{Initialize Wall Timer with period} \(\delta t\) 

\While{Wall Timer Callback}
    \State \(\textit{pose} \gets \text{getVehiclePose()}\)
    \State \(\textit{vMax} \gets \text{checkRemoteControl()}\) \Comment{Determine max velocity based on raceline, remote control, or flags} 
    \If {\(\text{timeoutDetected()}\)}
        \State \(\textit{vMax} \gets 0\)  \Comment{Enforce stop if timeout occurs}
    \EndIf    
    \State \(\textit{nearestIdx} \gets \text{newtonRaphson}(\textit{pose}, \textit{racelines}[active])\)
      \State \(\textit{path} \gets \text{buildPath}(\textit{racelines}[active], \textit{nearestIdx}, \textit{vMax})\)
    \State \(\textit{localPath} \gets \text{globalToLocal}(\textit{path}, \textit{pose})\)
    \State \textbf{publish}(\(\textit{localPath}\))
\EndWhile

\end{algorithmic}
\end{algorithm}
\vspace{-.1in}

\begin{algorithm}[H]
\caption{Vehicle Control (Essential Pseudocode)}
\label{alg:vehicle_control}
\begin{algorithmic}[1]
\While{vehicle is active}
  \State $\textit{joystickCmd} \gets \text{checkJoystick()}$ \Comment{Check if joystick override button is pressed}
  \If{$\text{timeoutDetected()}$ \text{in dependency modules}}
    \State $\textit{throttleCmd} \gets 0,\; \textit{brakeCmd} \gets \text{max allowed},\; \textit{steeringCmd} \gets \text{last value}$
    \State \textbf{continue} 
  \EndIf

  \If{$\textit{joystickCmd}.\text{overrideActive}$}
    \State $\textit{throttleCmd} \gets 0$ 
    \State $\textit{brakeCmd} \gets \textit{joystickCmd}.\text{brake}$
    \State $\textit{steeringCmd} \gets \textit{joystickCmd}.\text{steering}$
  \Else
  \Comment{Longitudinal Control}

    \State $v_{\text{ref}} \gets \text{localTrajectory}[2].v$ 
    \State $\Delta v \gets (v_{\text{ref}} - v_{\text{car}})$
    \If{$\Delta v > \textit{throttleDeadband}$}
       \State $\textit{throttleCmd} \gets \text{PIDthrottle}(\Delta v)$
       \State $\textit{brakeCmd} \gets 0$
    \ElsIf{$\Delta v < -\textit{brakeDeadband}$}
       \State $\textit{brakeCmd} \gets \text{PIDbrake}(\Delta v)$
       \State $\textit{throttleCmd} \gets 0$
    \Else
       \State $\textit{throttleCmd} \gets 0,\; \textit{brakeCmd} \gets 0$ \Comment{Engine Braking}
    \EndIf
    \State \text{checkEngineRPM}($\text{rpm}, \textit{gearCmd}$) 

    \Comment{Lateral Control}
    \State $L_d \gets \min\bigl(L_{\mathrm{d,min}} + k \cdot v_{\text{car}},\, L_{\mathrm{d,max}}\bigr)$ 
    \State \textit{lookAheadPoint} $\gets$ \text{findLookaheadPoint}(\text{path})
    \State $\alpha \gets \text{computeHeadingError}(\textit{pose}, \textit{lookAheadPoint})$
    
    \State \textit{lookahead\_angle} $\gets \tan^{-1} \bigl(\frac{\textit{lookAheadPoint.y}}{\textit{lookAheadPoint.x}}\bigr)$
    
    \State $\delta \gets \tan^{-1} \biggl(\frac{2 \cdot L \cdot \sin(\textit{lookahead\_angle})}{L_d} \biggr)$
    \vspace{0.08in}
    \State $\delta \gets \text{clamp}(\delta, \delta_{\min}, \delta_{\max})$

    \State $\textit{steeringCmd} \gets \delta \times \textit{steeringRatio}$

  \EndIf

    \State \textbf{sendLongC}($\textit{throttleCmd}, \textit{brakeCmd}, \textit{gearCmd}$)
    \State \textbf{sendLateralCommand}($\textit{steeringCmd}$)
    
\EndWhile
\end{algorithmic}
\end{algorithm}

\section*{\textbf{Supplementary Material}: Pseudocode, parameters, ROS Messages, and System Failure}
\subsection{Parameters}

\begin{table}[H]
    \centering
    \renewcommand{\arraystretch}{1.3}
    \begin{tabular}{ l c c l c c }
        \toprule
        \textbf{Parameter} & \textbf{Value} & \textbf{Unit} & \textbf{Parameter} & \textbf{Value} & \textbf{Unit} \\
        \midrule
        \multicolumn{6}{l}{\textbf{Vehicle Model}} \\
        Wheelbase & 2.9718 & m & Steering to Road Wheel Ratio & 15.0 & - \\
        Maximum Steering Angle & 230.0 & degrees & Maximum Throttle Command & 55.0 & \% \\
        Maximum Brake Command & 1800.0 & kPa & & & \\
        \midrule
        \multicolumn{6}{l}{\textbf{Control Node Parameters}} \\
        Control Loop Frequency & 50.0 & Hz & Lateral Error Threshold & 3.5 & m \\
        Trajectory Timeout & 0.2 & s & Joystick Timeout & 5.0 & s \\
        Throttle Deadband & 0.2 & m/s & Brake Deadband & 0.4 & \% \\
        \midrule
        \multicolumn{6}{l}{\textbf{PID Brake Controller}} \\
        Proportional Gain ($K_p$) & 300.0 & - & Integral Gain ($K_i$) & 0.0 & - \\
        Derivative Gain ($K_d$) & 2.0 & - & Command Max & 1800.0 & kPa \\
        Integral Max & 15.0 & - & & & \\
        \midrule
        \multicolumn{6}{l}{\textbf{PID Throttle Controller}} \\
        Proportional Gain ($K_p$) & 17.0 & - & Integral Gain ($K_i$) & 16.0 & - \\
        Derivative Gain ($K_d$) & 1.1 & - & Command Max & 55.0 & \% \\
        Integral Max & 0.5 & - & & & \\
        \midrule
        \multicolumn{6}{l}{\textbf{Pure Pursuit Controller}} \\
        Steering Gain ($K_p$) & 1.0 & - & Lookahead Distance Ratio & 0.63 & - \\
        Minimum Lookahead Distance & 15.0 & m & Maximum Lookahead Distance & 27.0 & m \\
        \midrule
        \multicolumn{6}{l}{\textbf{Gear Shift Parameters}} \\
        Shift Up (1st Gear) & 4000.0 & RPM & Shift Down (2nd Gear) & 2000.0 & RPM \\
        Shift Up (2nd Gear) & 4200.0 & RPM & Shift Down (3rd Gear) & 2100.0 & RPM \\
        Shift Up (3rd Gear) & 4300.0 & RPM & Shift Down (4th Gear) & 2200.0 & RPM \\
        Shift Up (4th Gear) & 4400.0 & RPM & Shift Down (5th Gear) & 2300.0 & RPM \\
        Shift Up (5th Gear) & 4500.0 & RPM & Shift Down (6th Gear) & 2400.0 & RPM \\
        Shift Time & 500 & ms & & & \\
        \midrule
        \multicolumn{6}{l}{\textbf{Planner Node Parameters}} \\
        Planner Rate & 50 & Hz & Path Step Time & 0.05 & s \\
        Path Duration & 2.5 & s & Localization Timeout & 0.2 & s \\
        Remote Control Timeout & 5.0 & s & Remote Control Rate & 20 & Hz \\
        \bottomrule
    \end{tabular}
    \caption{Vehicle and Control Parameters}
    \label{tab:vehicle_params}
\end{table}

\newpage
\section*{\textbf{Supplementary Material}: Pseudocode, parameters, ROS Messages, and System Failure}
\subsection{ROS Messages}

\begin{table}[H]
    \centering
    \renewcommand{\arraystretch}{1.2}
    \begin{tabular}{|l|p{7cm}|l|l|}
        \hline
        \textbf{Message Type} & \textbf{Description} &\textbf{Field Name} & \textbf{Data Type} \\ 
        \hline
        \multirow{12}{*}{Basestation.msg} 
          & \multirow{12}{*}{\makecell[l]{Published from the base-station to the autonomy stack online.}}
          & stamp & builtin\_interfaces/Time \\ \cline{3-4}
          &  & v\_max & float32 \\ \cline{3-4}
          &  & raceline\_index & int8 \\ \cline{3-4}
          &  & veh\_flag & int8 \\ \cline{3-4}
          &  & track\_flag & int8 \\ \cline{3-4}
          &  & enable\_engine & bool \\ \cline{3-4}
          &  & enable\_driving & bool \\ \cline{3-4}
          &  & enable\_joystick\_control & bool \\ \cline{3-4}
          &  & target\_velocity & float32 \\ \cline{3-4}
          &  & steering\_cmd & float32 \\ \cline{3-4}
          &  & brake\_amount & float32 \\ \cline{3-4}
          &  & throttle\_lockout & bool \\ \hline
        \multirow{30}{*}{Dashboard.msg} 
        & \multirow{30}{*}{\makecell[l]{Published from the autonomy stack to the base-station. \\Used in real-time to monitor the performance.}}
        & stamp & builtin\_interfaces/Time \\ \cline{3-4}
        & &cmd\_gear & int8 \\ \cline{3-4}
        & &actual\_gear & int8 \\ \cline{3-4}
        & &cmd\_throttle & int8 \\ \cline{3-4}
        & &actual\_throttle & int8 \\ \cline{3-4}
        & &cmd\_brake & int16 \\ \cline{3-4}
        & &actual\_brake\_front & int16 \\ \cline{3-4}
        & &actual\_brake\_rear & int16 \\ \cline{3-4}
        & &cmd\_steering\_degree & int16 \\ \cline{3-4}
        & &actual\_steering\_degree & int16 \\ \cline{3-4}
        & &heading\_error & float32 \\ \cline{3-4}
        & &cross\_track\_error & float32 \\ \cline{3-4}
        & &velocity\_error & float32 \\ \cline{3-4}
        & &target\_velocity\_mps & float32 \\ \cline{3-4}
        & &actual\_velocity\_mps & float32 \\ \cline{3-4}
        & &purepursuit\_lookahead\_distance & float32 \\ \cline{3-4}
        & & purepursuit\_lookahead\_angle\_rad & float32 \\ \cline{3-4}
        & &position\_x & float32 \\ \cline{3-4}
        & &position\_y & float32 \\ \cline{3-4}
        & &position\_z & float32 \\ \cline{3-4}
        & &position\_r & float32 \\ \cline{3-4}
        & &position\_p & float32 \\ \cline{3-4}
        & &position\_yaw & float32 \\ \cline{3-4}
        & &velocity\_x & float32 \\ \cline{3-4}
        & &velocity\_y & float32 \\ \cline{3-4}
        & &velocity\_z & float32 \\ \cline{3-4}
        & &trust & float32 \\ \cline{3-4}
        & &status & int8 \\ \cline{3-4}
        & &engine\_speed\_rpm & float32 \\ \cline{3-4}
        & &vehicle\_speed\_kmph & float32 \\ \hline
        \multirow{9}{*}{State.msg} 
        & \multirow{9}{*}{\makecell[l]{The output of the state-estimation node. \\ Used mainly by planning and control modules.}}
        & header & std\_msgs/Header \\ \cline{3-4}
        & &position & geometry\_msgs/Point \\ \cline{3-4}
        & &rpy & geometry\_msgs/Point \\ \cline{3-4}
        & &velocity & geometry\_msgs/Point \\ \cline{3-4}
        & &angular\_velocity & geometry\_msgs/Point \\ \cline{3-4}
        & &slip\_angle & geometry\_msgs/Point \\ \cline{3-4}
        & &slip\_angle\_front & float64 \\ \cline{3-4}
        & &slip\_angle\_rear & float64 \\ \cline{3-4}
        & &trust & float64 \\ \cline{3-4}
        & &status & int8 \\ \hline
        \multirow{4}{*}{Path.msg} 
        & \multirow{4}{*}{The output of the planner. Used by the controller module.} 
        & header & std\_msgs/Header \\ \cline{3-4}
        & &current\_state & State \\ \cline{3-4}
        & &path & State[] \\ \cline{3-4}
        & &status & int8 \\ \hline
    \end{tabular}
    \caption{Essential ROS Messages Fields}
    \label{tab:ros_messages}
\end{table}

\section*{\textbf{Supplementary Material}: Pseudocode, parameters, ROS Messages, and System Failure}
\subsection{System Failure}

\begin{figure}[h]
    \centering
    \includegraphics[scale=0.38]{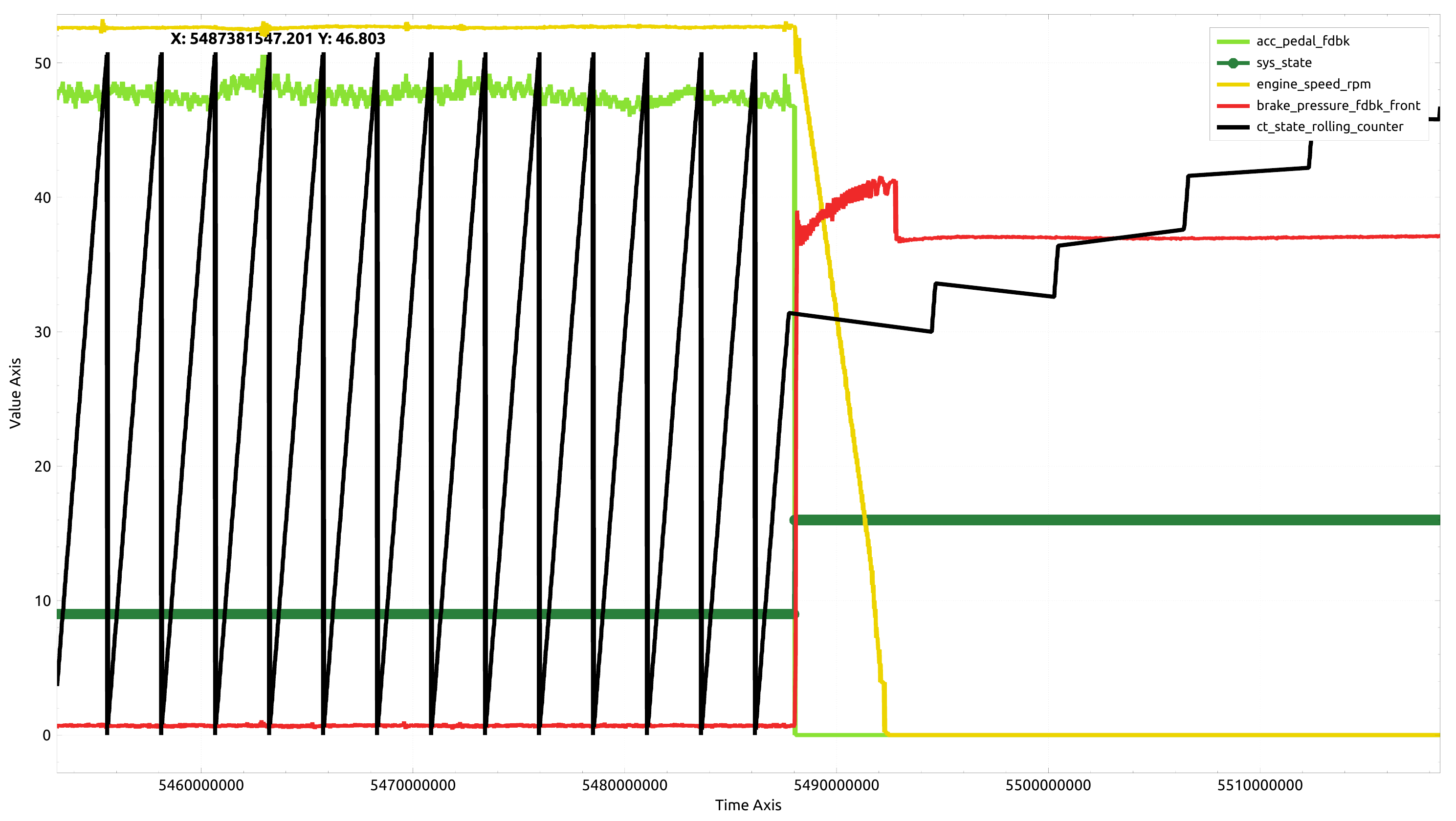}
    \caption{\small CAN logs corresponding to the car crash. The figure displays the rolling counter (black), system state (dark green), throttle percentage (light green), brake pressure in Newtons [x$50$] (red), and engine RPM [x$10^2$](yellow).}
    \label{fig:crash_can}
    \vspace{-0.25in}
\end{figure}

Figure~\ref{fig:crash_can} presents the CAN logs corresponding to the crash incident that occurred in September 2025 at IMS. As explained in Section V, the ROS framework froze, which delayed the transmission of the rolling counter to the raptor. Consequently, the lower-level system received incorrect rolling counter values (depicted in black), thereby initiating an emergency shutdown. This event is characterized by a transition in the system state (shown in dark green) from normal driving to emergency shutdown. As a result, the throttle position (shown in light green) immediately decreased to zero, while the brake pressure (shown in red) surged to a maximum of 1800 N, causing the engine RPM (depicted in yellow) to fall to zero.

\end{document}